\documentclass{article}
\usepackage{amssymb}

\usepackage{array}
\usepackage{tabularx}
\newcolumntype{Y}{>{\centering\arraybackslash}X}

\usepackage[preprint]{corl_2025} 
\newcommand{\ours}{HALO}

\title{\ours{}: Human Preference Aligned Offline Reward Learning for Robot Navigation}

\usepackage{xcolor}
\usepackage{amssymb}
\usepackage{graphicx}
\usepackage{mathtools}
\usepackage{amsmath}
\usepackage{gensymb}
\usepackage{float}
\usepackage{multirow}
\usepackage{makecell}
\usepackage{mathtools}
\usepackage{hyperref}
\usepackage{booktabs} 
\hypersetup{
    colorlinks=true,
    linkcolor=blue,
    filecolor=magenta,      
    urlcolor=blue,
    pdftitle={Overleaf Example},
    pdfpagemode=FullScreen,
    }

\usepackage{tabularx}
    \newcolumntype{L}{>{\raggedright\arraybackslash}X}
\usepackage[utf8]{inputenc}
\usepackage[english]{babel}

\usepackage{amsthm}
\usepackage{fancyhdr}

\usepackage{subcaption}

\author{
\parbox{\textwidth}{\centering
Gershom Seneviratne \quad
Jianyu An \quad
Sahire Ellahy \quad
Kasun Weerakoon \\
Mohamed Bashir Elnoor \quad
Jonathan Deepak Kannan \quad
Amogha Thalihalla Sunil \\
Dinesh Manocha
} \\ \\
University of Maryland, College Park
}

\fancypagestyle{fancyfirstpage}{
  \fancyhf{}  
  \fancyfoot[L]{\footnotesize 9th Conference on Robot Learning (CoRL 2025), Seoul, Korea.}
  
}

\begin{document}
\maketitle


\begin{abstract}
In this paper, we introduce \ours{}, a novel Offline Reward Learning algorithm that quantifies human intuition in navigation into a vision-based reward function for robot navigation. \ours{} learns a reward model from offline data, leveraging expert trajectories collected from mobile robots. During training, actions are uniformly sampled around a reference action and ranked using preference scores derived from a Boltzmann distribution centered on the preferred action, and shaped based on binary user feedback to intuitive navigation queries. The reward model is trained via the Plackett-Luce loss to align with these ranked preferences. To demonstrate the effectiveness of HALO, we deploy its reward model in two downstream applications: (i) an offline learned policy trained directly on the HALO-derived rewards, and (ii) a model-predictive-control (MPC) based planner that incorporates the HALO reward as an additional cost term. This showcases the versatility of HALO across both learning-based and classical navigation frameworks. Our real-world deployments on a Clearpath Husky across diverse scenarios demonstrate that policies trained with HALO generalize effectively to unseen environments and hardware setups not present in the training data. \ours{} outperforms state-of-the-art vision based navigation methods, achieving at least a 33.3\% improvement in success rate, a 12.9\% reduction in normalized trajectory length, and a 26.6\% reduction in Fréchet distance compared to human expert trajectories.

\end{abstract}


\let\thefootnote\relax
\footnotetext{\tiny Video, Code and more information: \url{http://gamma.umd.edu/halo/}}


\section{Introduction} \label{sec:intro}

Autonomous visual navigation is a fundamental capability for mobile robots operating in complex, real-world environments \cite{Paluch2020, raj2024rethinking, weerakoon2024behav, shah2023vint}. Traditional systems often rely on depth sensors such as LiDAR or stereo cameras to estimate geometry and avoid obstacles\cite{}. While effective, these sensors introduce significant cost, power, and hardware complexity, limiting scalability across robot platforms\cite{weerakoon2024behav, seneviratne2024cross}. In contrast, RGB cameras offer a low-cost and widely deployable sensing modality. However, building reliable navigation systems based solely on RGB input remains challenging—particularly due to the difficulty of inferring navigability and obstacle relevance directly from raw visual observations under dynamic lighting, ambiguous pathways, or unstructured terrain. In this context, a method that can leverage human intuition to interpret visual scenes and guide navigation decisions could enhance trajectory safety and improve navigation success \cite{payandeh2024social, shah2022offline}.

Reinforcement Learning (RL) has demonstrated significant potential for navigation, allowing robots to learn from direct interaction with the environment (i.e., online RL) and adapt based on experiential data \cite{tung2018socially, sigal2023improving}. Many online RL methods utilize simulators during training, leveraging parallel processing and faster-than-real-time learning to improve efficiency. However, a key limitation is the \textit{sim-to-real transfer gap} \cite{yao2024sonic}, where policies trained in simulation often fail to generalize effectively to real-world environments. To mitigate this, recent works have leveraged large-scale navigation datasets \cite{jiang2021rellis,karnan2022scand} to train policies offline—an approach known as offline RL—which enables learning directly from real-world observations without requiring costly or unsafe online interaction \cite{shah2022offline, weerakoon2023vapor, caesar2020nuscenes, geyer2020a2d2}. Advances in offline RL \cite{kostrikov2021offline, kumar2020conservative} have further enabled robots to learn robust policies in a data-efficient manner using pre-collected experience, reducing reliance on online interactions. Despite these advances, both online and offline RL approaches typically depend on \textit{hand-engineered reward functions}, requiring substantial domain expertise to identify key components and mathematically formalize human intuition \cite{kapoor2023socnavgym, liang2021crowd}. Moreover, reward functions must be meticulously tuned, often necessitating multiple rounds of real-world training and testing, a costly and resource-intensive process. Furthermore, many RL-based navigation algorithms rely on LiDAR or depth cameras for obstacle distance estimation during reward computation, which, while effective, introduces scalability challenges due to sensor cost and deployment constraints \cite{patel2021dwa, weerakoon2023vapor}.


Recent research has also explored the use of large language models (LLMs) and vision-language models (VLMs) to tackle the challenges of visual navigation. These models offer an alternative paradigm: instead of explicitly optimizing reward functions, they use natural language understanding to interpret goals, identify obstacles, or infer human intent from egocentric observations \cite{mondorf2024beyond, huang2022towards, parmar2024logicbench}. Building on this foundation, recent research has explored the application of these models to enhance autonomous robot navigation \cite{song2024vlm, Huang2022VLMaps, Sathyamoorthy2024CoNVOICN, elnoor2024robot}. Typically, these approaches involve providing the model with an egocentric view of the environment, along with a detailed text description of the scene, and then prompting the model to generate a high-level plan or to infer the likely behavior of nearby agents. Although these methods offer promising avenues for capturing nuanced human intentions and facilitating adaptive navigation strategies, they have some limitations. Specifically, LLM/VLM-based solutions often demand significant computational resources, require continuous internet connectivity due to their reliance on cloud-based services, and are not easily deployable on edge hardware \cite{kumar2020conservative, kostrikov2021offline}. Consequently, these constraints impede their real-time performance and scalability, ultimately limiting their applicability in safety-critical and resource-constrained navigation scenarios.

\textbf{Main contributions:} To address the limitations of hand-crafted reward functions in vision based navigation, we propose \textbf{HALO}—\textbf{H}uman Preference \textbf{AL}igned \textbf{O}ffline Reward Learning—a novel framework that uses human feedback to train reward models from egocentric visual inputs and action-conditioned trajectories. Our key contributions are:

\begin{itemize}

    \item \textbf{A preference-driven reward learning framework} that uses Plackett-Luce loss to align ranked action sets with human visual intuition, based on responses to binary feasibility queries (e.g., "Can the robot turn left?", "Can the robot accelerate?") from egocentric camera views. The resulting reward model is deployed in two downstream settings: (i) training a goal-conditioned offline policy, and (ii) augmenting a model predictive control (MPC) planner with the learned reward as a cost term—demonstrating strong generalization across diverse indoor and outdoor environments in both learning-based and classical navigation frameworks. Furthermore, the annotated user preference dataset for sub-optimal trajectories will be released along with the code for the paper.

    \item \textbf{An action-conditioned visual feature aggregation mechanism} that uses a binary mask of the predicted robot path, generated via homography, to identify spatially relevant regions in the image. This mask is processed through a CNN to produce a weighting map that modulates visual features, allowing the model to aggregate highly relevant image information based on the intended action.

    \item \textbf{Extensive real-world evaluation} on the Clearpath Husky platform, showcasing HALO’s ability to generalize across diverse indoor and outdoor environments not encountered during training. Our reward-model-based policies operate in real time at approximately 50 Hz on a laptop (Nvidia 3060 GPU, Intel Core i7 CPU), and consistently outperforms state of the art vision based navigation methods rewards—achieving an 33.3\% improvement in success rate, 12.9\% reduction in normalized trajectory length, and 26.6\% lower Fréchet distance to expert trajectories.
    
\end{itemize}

\section{Related Work} \label{sec:related_work}

\begin{figure*}[t]
    \label{fig:reward_model_architecture}
      \centering  
      \includegraphics[width=\columnwidth]{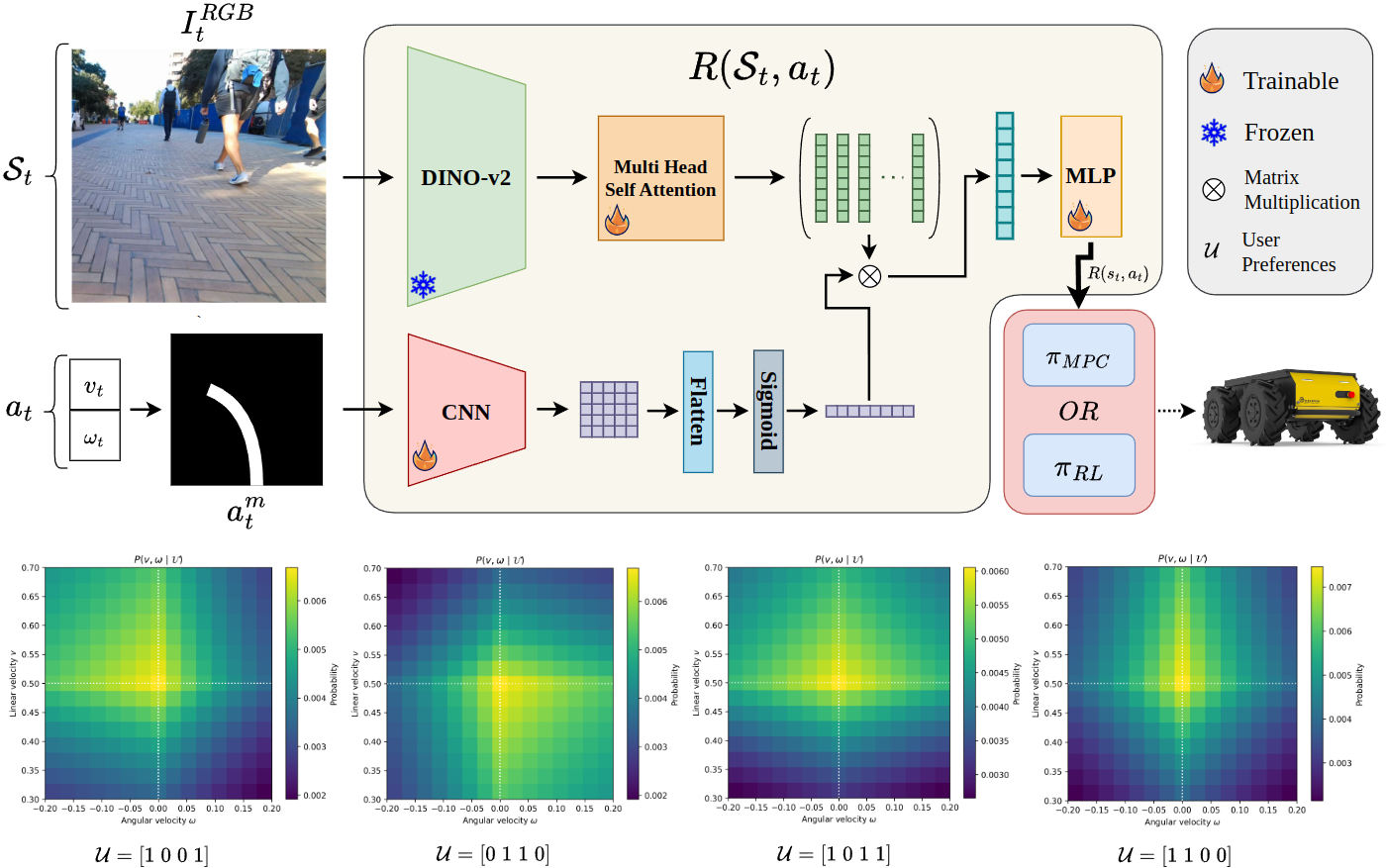}
      \caption {\small{Architecture of the proposed reward model. Given the current observation \(I_t^{RGB}\), a frozen DINO-v2 encoder extracts patch embeddings. Simultaneously, the candidate action \(a_t = (v_t, \omega_t)\) is projected into image space via a homography transform to produce a trajectory mask \(a_t^m\), which is passed through a trainable CNN to yield a spatial relevance weighting. This vector modulates the image patch embeddings, focusing on regions relevant to the trajectory. The resulting feature is passed through a trainable MLP to produce the scalar reward \(R(s_t, a_t)\). This reward can be used in either a model predictive controller (\(\pi_{\text{MPC}}\)) or an offline reinforcement learning policy (\(\pi_{\text{RL}}\)). \textbf{Bottom:} Boltzmann distributions \(P(v, \omega \mid \mathcal{U})\) generated from binary user preferences \(\mathcal{U}\), used to assign preference scores to actions during reward model training.}}        
      \label{fig:reward_model_architecture}
      \vspace{-10pt}
\end{figure*}

In this section, we review key literature on autonomous visual navigation, reward modeling based on preference data, and the use of LLMs/VLMs for social navigation.

\subsection{Navigation Algorithms for Mobile Robots}

Classical geometric and rule-based methods laid the foundation for autonomous navigation \cite{thrunyz1997dynamic}, but their reliance on depth sensors such as LiDAR limits scalability due to high cost and hardware complexity \cite{kulathunga2024resilient, weerakoon2024behav}. Reinforcement learning (RL) methods enable robots to learn adaptive navigation behaviors through interaction with their environments, often in simulation \cite{dosovitskiy2017carla}. Building on this, offline RL allows policies to be learned directly from large, real-world datasets without requiring online exploration \cite{shah2022offline, weerakoon2023vapor}. However, many offline RL approaches still depend on hand-crafted reward functions, which are difficult to design for nuanced tasks like obstacle avoidance or visual goal-following, and often require extensive tuning \cite{kim2024transformable}. For example, Shah et al. use multiple manually defined objectives to train multi-objective policies \cite{shah2022offline}, but none of these explicitly account for obstacle avoidance, limiting their scalability and generalization to complex, real-world scenarios.

\subsection{Reward Modeling Based on Preference Data}

Preference-based reward modeling has emerged as an effective way to align reinforcement learning agents with human intentions. Christiano et al. introduced a method using human preferences over trajectory segments to train reward models for policy optimization \cite{christiano2017deep}. Lee et al. extended this with PEBBLE, improving sample efficiency via unsupervised pre-training and experience relabeling \cite{lee2021pebble}. More recently, Wang et al. leveraged VLMs to automate preference labeling, eliminating manual annotation for training reward models in simulation \cite{Wang2024RLVLMFRL}. Most prior methods rely on the Bradley-Terry model \cite{BradleyTerry1952RankAO}, which simplifies feedback into binary comparisons, limiting nuance and introducing noise into the reward signal.

\subsection{Usage of VLMs and LLMs for Navigation}
Large Language Models (LLMs) and Vision-Language Models (VLMs) have shown strong reasoning and semantic understanding capabilities, making them promising for navigation tasks \cite{Zhai2024FineTuningLV}. Huang et al. \cite{Huang2022VLMaps} and Zhang et al. \cite{Zhang2024NaVidVV} applied VLMs for instruction-following in indoor navigation, while Song et al. \cite{song2025vltgstrajectorygenerationselection} and Sathyamoorthy et al. \cite{Sathyamoorthy2024CoNVOICN} extended this to outdoor environments. These systems typically use LLMs/VLMs to generate high-level plans, leaving low-level execution to traditional planners. However, they often suffer from hallucinations, high computational demands, and cloud latency, making them less suitable for real-time, edge-based deployment.



\section{Background}\label{sec:background}

In this section, we introduce key preliminaries necessary for a deeper exploration of the topics discussed in this work.



\subsection{Markov Decision Processes}

Robot navigation can be modeled as a Markov Decision Process (MDP) with continuous states and actions, defined by the tuple \((\mathcal{S}, \mathcal{A}, \mathcal{T}, \mathcal{R}, \gamma)\). Here, \(\mathcal{S}\) is the state space representing all possible environment configurations, and \(\mathcal{A}\) is the set of available robot actions. The transition function \(\mathcal{T}(s'|s,a)\) defines the probability of reaching state \(s'\) from \(s\) after action \(a\). The reward function \(\mathcal{R}(s,a)\) assigns an immediate reward for executing action \(a\) in state \(s\). The discount factor \(\gamma\) balances the importance of immediate versus future rewards, shaping long-term navigation behavior.

In this formulation, a policy \(\pi: \mathcal{S} \rightarrow \mathcal{P}(\mathcal{A})\) maps each state to a distribution over continuous actions. The optimal policy \(\pi^*\) is the one that maximizes the expected cumulative reward over trajectories, guiding effective robot navigation.


\subsection{Offline Reinforcement Learning}

Offline Reinforcement Learning (Offline RL) enables policy learning from a fixed dataset without further environment interaction, making it valuable in scenarios where exploration is costly or unsafe \cite{levine2020offline}. Key aspects of Offline RL include:  

Notable algorithms, such as Conservative Q-Learning (CQL) \cite{kumar2020conservative} and Implicit Q-Learning \cite{kostrikov2021offline}, tackle challenges like distribution shift (mismatch between dataset and learned policy), extrapolation error (Q-value overestimation for unseen states), and bootstrapping error (propagation of value estimation inaccuracies).

\subsection{Plackett-Luce Model for preference modeling}
\label{sec:PLM}
The Plackett-Luce model is a probabilistic framework for preference modeling \cite{PlackettL1975TheAO}
. Given a set of actions \(\mathcal{A} = \{a_1, a_2, \dots, a_n\}\) and associated scores \(\theta_i\), the probability of observing a specific ranking \(\sigma\) is defined as

\begin{equation}
P(\sigma) = \prod_{j=1}^{n} \frac{\exp(\theta_{\sigma(j)})}{\sum_{k=j}^{n} \exp(\theta_{\sigma(k)})}.
\label{eq:plackett_luce}
\end{equation}

This formulation converts the scores into a probability distribution over rankings by exponentiating the scores and normalizing them sequentially.

\section{Our Approach: \ours{}}

We present HALO, an offline algorithm that uses the Plackett-Luce framework to learn a reward model aligned with human navigational preferences. In this section, we will focus on six key aspects of our approach:  (1) the dataset used for training and evaluation, (2) formulating the reward learning problem, (3) generating probabilistic scores for actions based on human feedback, (4) offline reward learning based on the Plackett-Luce model, (5) the proposed reward model architecture, and (6) robot navigation using the learned reward model.

\subsection{Dataset}
\label{sec:Dataset}

We use the Socially CompliAnt Navigation Dataset (SCAND) \cite{karnan2022scand} for offline training. It includes demonstrations in indoor and outdoor settings, where a human operator teleoperates a wheeled or legged robot around the UT Austin campus. From the 139 available scenes, we manually annotated 25 scenes (\(\approx\)107,000 frames) with human preference scores (Section~\ref{sec:ProbScores}).

To augment this, we collected 116 additional scenes (\(\approx\)33,000 frames) using a legged robot.  Approximately 100 of these consist of short, truncated trajectories that would result in collisions or unsafe behavior if executed further. These were deliberately included as negative examples to counterbalance the overwhelmingly successful demonstrations in SCAND. Section~\ref{sec:ProbScores} explains how these negative trajectories are incorporated into the preference scoring framework to guide the reward model toward safer behavior and improve generalization in safety-critical situations.

\subsection{Formulating the Reward Learning Problem}
\label{sec:FormulatingRM}

We model human-aligned navigation as an MDP and focus on learning a reward function $\mathcal{R}(s_t, a_t)$ that captures human navigational preferences from offline data.

In our framework, the action space $\mathcal{A}$ consists of continuous control commands represented by linear and angular velocity pairs $(v, \omega)$, and the state space $\mathcal{S}$ consists of the robot’s egocentric RGB camera observation $I^{RGB}$. The reward model is goal-independent and quantifies the quality of actions based solely on the egocentric visual input, guided by human preference data.

The policy $\pi(a_t | s_t)$ is defined as a stochastic distribution over continuous actions conditioned on the current state. This formulation supports smooth and adaptive control strategies, suitable for navigation in dynamic and complex environments.

\subsection{Preference Scores from Human Feedback}
\label{sec:ProbScores}

To capture human intuition, we collect binary responses to five navigation queries based on the robot’s egocentric view and expert action \((v^*, \omega^*)\): \textit{(1) Can the robot turn left? (2) Turn right? (3) Decelerate? (4) Accelerate? (5) Is the robot in immediate danger or behaving suboptimally?} These responses assess the feasibility of alternative actions. To reduce annotation effort, annotators were able to reuse previous responses by simply advancing through frames without re-entering inputs, as long as the answers to the navigation queries remained unchanged. Labeling resumed when a meaningful change in the scene—such as a hallway turn or a pedestrian entering—required a different response, based on the annotator’s visual judgment rather than a predefined threshold.

If the user flags danger or suboptimal behavior, a corrective reference action \((v^*, \omega^*)\) is assigned based on user preference data. This differs from the expert action recorded in the dataset and typically involves reduced linear velocity and a sharp turn. This encourages the reward model to favor safer and more responsive behaviors in similar scenarios.

We generate a discrete, dynamically feasible action set centered around \((v^*, \omega^*)\) that spans the robot’s field of view:

\[
\mathcal{A}_{\text{local}} = \{(v, \omega) \mid v \in \mathcal{V},\ \omega \in \Omega \}
\]

Each action \((v, \omega) \in \mathcal{A}_{\text{local}}\) is assigned a probability score based on its proximity to \((v^*, \omega^*)\), using a separable Boltzmann distribution (Eq. \ref{eq:sep}, \ref{eq:boltz}):

\begin{equation}
\label{eq:sep}
P(v, \omega, \mathcal{U}) = P(v, \mathcal{U}) \cdot P(\omega, \mathcal{U}),
\end{equation}

\begin{equation}
\label{eq:boltz}
P(v, \mathcal{U}) = \frac{\exp\left(-\frac{|v - v^*|}{\tau_v(\mathcal{U})} \right)}
{\sum\limits_{v' \in \mathcal{V}} \exp\left(-\frac{|v' - v^*|}{\tau_v(\mathcal{U})} \right)},
\quad
P(\omega, \mathcal{U}) = \frac{\exp\left(-\frac{|\omega - \omega^*|}{\tau_\omega(\mathcal{U})} \right)}
{\sum\limits_{\omega' \in \Omega} \exp\left(-\frac{|\omega' - \omega^*|}{\tau_\omega(\mathcal{U})} \right)}.
\end{equation}

The temperature values \(\tau_v(\mathcal{U})\) and \(\tau_\omega(\mathcal{U})\) are selected adaptively based on user preferences \(\mathcal{U} \in \{0,1\}^4\), which correspond to the four directional queries. When a directional preference is indicated, the corresponding temperature is set such that the preferred direction receives at least 95\% of the total probability mass in its respective marginal distribution. In the absence of preference, the distribution is flattened by assigning a higher temperature. This ensures that undesirable actions are assigned a selection probability of less than 0.05, and that the distribution sharpens or flattens depending on the user's certainty.

To reflect the user’s perceived desirability of a scene, the final score is scaled by a scalar factor \(\lambda\) (Eq. \ref{eq:lambda}), defined as:

\begin{equation}
\label{eq:lambda}
\lambda(\mathcal{U}, \mathcal{U}_{\text{danger}}) =
\begin{cases}
- \dfrac{1}{1 + \sum_{i=1}^4 \mathcal{U}_i}, & \text{if } \mathcal{U}_{\text{danger}} = 1, \\[0pt]\\
1 + \sum_{i=1}^4 \mathcal{U}_i, & \text{otherwise}.
\end{cases}
\end{equation}

The final preference score assigned to an action \((v, \omega)\) is therefore:

\begin{equation}
\text{Pref}(v, \omega \mid \mathcal{U}) = \lambda(\mathcal{U}, \mathcal{U}_{\text{danger}}) \cdot P(v, \omega \mid \mathcal{U}).
\end{equation}

These scores are used to rank the \(n_v \times n_\omega\) candidate actions, with the reference action always ranked highest. The resulting preference-ordered list \(\sigma\) serves as supervision for reward model training.

\subsection{ Reward model learning based on preference data using the Plackett Luce loss}
\label{}

We use the Plackett-Luce model Section (\hyperref[sec:PLM]{3.3}) to learn a reward function that maximizes the probability of observing the preference rankings generated in Section \hyperref[sec:ProbScores]{4.2}. To achieve this, we minimize the negative log-likelihood of the Plackett-Luce ranking probability defined in Eq. \eqref{eq:plackett_luce}, ensuring that the learned reward function aligns with human preferences.

\begin{equation}
\mathcal{L}_{\text{PL}} = - \sum_{i=1}^{n-1} \theta_{\sigma(i)} + \sum_{i=1}^{n-1} \log \left( \sum_{j=i}^{n} \exp(\theta_{\sigma(j)}) \right)
\label{eq:plackett_luce_loss}
\end{equation}

By minimizing the Plackett-Luce loss defined in Eq. \eqref{eq:plackett_luce_loss}, the reward model is trained to assign higher values to actions that appear earlier in the ranked preference list, ensuring that the learned reward function captures human intuition in navigation. This loss offers a scalable and context-aware formulation for learning from ranked actions, capturing the relative importance of all candidates in a single ranking. Unlike pairwise comparison methods, it avoids the \(\binom{n_v \times n_\omega}{2}\) comparisons required per timestep, resulting in a more stable loss landscape and efficient optimization.

To the best of our knowledge, this is the first approach to incorporate Plackett-Luce preference-based reward learning for modeling human intuition in autonomous navigation. This formulation offers a structured, scalable, and globally consistent framework for aligning navigation policies with human preferences.

\subsection{Reward Model Architecture}

The overall architecture of our system is illustrated in Figure \hyperref[fig:reward_model_architecture]{1}. Given a state \(s_t = I^{RGB}\) and a candidate action \(a_t = (v_t, \omega_t) \in \mathbb{R}^2\), the model outputs a scalar reward \(\mathcal{R}(s_t, a_t)\).

\textbf{Image Feature Extraction:} We use DINOv2~\cite{oquab2023dinov2}, a self-supervised feature extractor based on the Vision Transformer (ViT) architecture, to extract patch-level embeddings from the input image \(I^{RGB}\). The frozen encoder outputs \(N_p\) spatial patch embeddings of dimension \(d_e\), denoted as \(F_t \in \mathbb{R}^{N_p \times d_e}\). These embeddings are further refined via \(N_{\text{sa}}\) Transformer layers with \(h\) self-attention heads, yielding \(F_t^{\text{sa}} \in \mathbb{R}^{N_p \times d_e}\).

\textbf{Action-Conditioned Visual Feature Aggregation:} Given an action \((v, \omega)\), we generate a binary trajectory mask by projecting the path the robot would follow under that action over a fixed time horizon onto the image plane using a homography transform (see Figure~\hyperref[fig:reward_model_architecture]{1}). This results in a binary mask \(a_t^m \in \mathbb{R}^{1 \times H_m \times W_m}\), which is processed by a lightweight CNN composed of stacked convolutional and pooling layers, followed by a final sigmoid activation. The output is a spatial relevance map \(W_t \in \mathbb{R}^{\sqrt{N_p} \times \sqrt{N_p}}\), which is flattened into an \(N_p\)-dimensional vector. This vector is then used to modulate the DINOv2 patch embeddings \(F_t^{\text{sa}} \in \mathbb{R}^{N_p \times d_e}\) via element-wise multiplication, allowing the model to emphasize image regions relevant to the given action and support action-conditioned perception analogous to attention-based gating.

\textbf{Final Reward Prediction:}  
The aggregated feature representation is passed through a simple MLP head that maps it directly to a scalar reward value. This architecture provides an efficient mechanism for capturing the relationship between visual input and action-specific context.

\subsection{ Training and Offline Navigation Policy}
Once trained, the reward model can be used in two ways: (i) to supervise a goal-conditioned offline policy, or (ii) to augment classical navigation algorithms such as Dynamic Window Approach (DWA) or Model Predictive Control (MPC) by serving as an additional cost term. 

To improve reward learning stability and generalization, we incorporate two regularization strategies during training: (i) a focal-style penalty applied to the raw preference scores generated in \hyperref[sec]{}, and (ii) a diversity regularizer that discourages similar rewards for dissimilar actions. These additions help the model produce more discriminative and informative reward outputs.

Additional training details, including offline policy learning methodology, MPC implementation and hyperparameters are provided in Appendix~\ref{appdx:impl_det}.

\section{Results and Analysis}

\subsection{Implementation and Comparisons}

We integrate our method into the Clearpath Husky wheeled robot for real-world experiments. 
The Clearpath Husky is outfitted with a Realsense d435i camera, and a laptop containing an Intel i7 processor and an NVIDIA RTX 3060 GPU. 
To compare against our reward model (section \ref{sec:FormulatingRM}), we also designed a hand-engineered reward (HER) model that accounted for distance to the goal, distance to any obstacles utilizing lidar data, and a smoothness term to discourage sudden acceleration. We validated our method by comparing the IQL \cite{kostrikov2021IQL} policy trained on our reward model to the IQL policy trained on hand-engineered rewards (HER). Additionally, we trained a behavioral cloning policy (BC) \cite{fujimoto2018addressing} on the same data used to train the reward model. We also compared our method with a classical method \cite{fox1997dynamic} and a learning-based vision-action method, VANP \cite{10802451}. 

\subsection{Evaluation Metrics}

We evaluate our reward model on three navigation metrics:  (1) Success rate, (2) Normalized trajectory length, and Fréchet distance \cite{alt1995computing} w.r.t. the human teleoperation. Detailed descriptions of these metrics and their computation can be found in Appendix~\ref{appdx:metrics}.

\subsection{Discussion}

\begin{figure}[t]
      \centering  
      \includegraphics[width=\columnwidth]{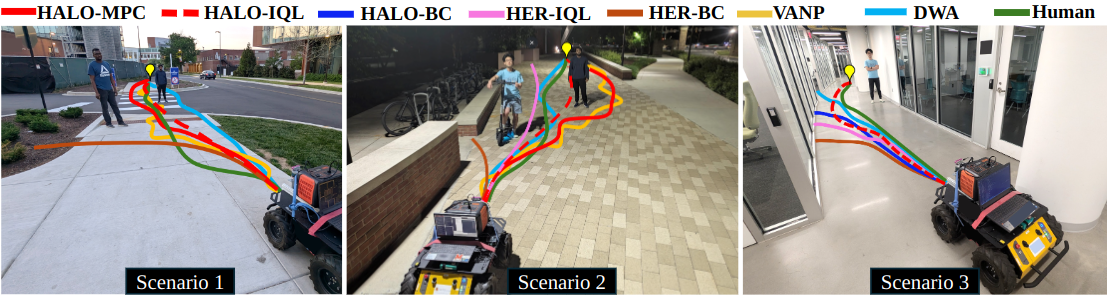}
      \caption{\small{ Comparison across three diverse real-world navigation scenarios (outdoors, low light, indoors). Each row shows the trajectory taken by different methods—DWA \cite{fox1997dynamic}, VANP \cite{10802451}, HER + IQL \cite{kostrikov2021IQL}/BC\cite{fujimoto2018addressing}, and HALO-based policies—highlighting differences in social compliance, obstacle avoidance, and goal-reaching behavior.}}
      \label{fig:qual}
\end{figure}

We present \ours{}’s performance results and comparisons quantitatively in Table~\ref{tab:comparisons} and qualitatively in Figure~\ref{fig:qual}. These comparisons span three diverse real-world navigation scenarios. In the following discussion, we analyze how each method performs within these settings, highlighting differences in social compliance and obstacle avoidance. \ours{} consistently outperforms baseline methods in terms of Success Rate and Fréchet Distance, demonstrating its ability to produce safer and more human-aligned trajectories without relying on LiDAR or hand-crafted rewards. Further qualitative results comparing HALO and HER methods are discussed in the Appendix \ref{appdx:qual}.

\begin{table}[H]
\centering
\setlength{\abovecaptionskip}{6pt}
\scriptsize
\begin{tabular}{|c|c|c|c|c|}
\hline
\textbf{Scenario} & \textbf{Method} & \textbf{Success Rate (\%) $\uparrow$} & \textbf{Normalized Traj. Length $\downarrow$} & \textbf{Fréchet Distance $\downarrow$ (m)} \\
\hline

\multirow{7}{*}{\shortstack{Scenario 1}} 
& DWA (\textdagger) \cite{fox1997dynamic} & 80 & 1.043 & 1.677 \\
& VANP \cite{10802451} & 60 & 1.205 & 1.216 \\
& HER + TD3-BC (\textdagger) \cite{fujimoto2021minimalist} & 0 & 0.420 & 12.651 \\
& HER + IQL (\textdagger) \cite{kostrikov2021IQL} & 50 & 1.118 & 3.665 \\
& \ours{} + TD3-BC & 10 & 0.565 & 6.758 \\
& \ours{} + IQL & 80 & 1.049 & 1.140 \\
& \ours{} + DWA & 70 & 1.212 & 0.892 \\
\hline

\multirow{6}{*}{\shortstack{Scenario 2}} 
& DWA (\textdagger) \cite{fox1997dynamic} & 100 & 1.157 & 0.678 \\
& VANP \cite{10802451} & 60 & 1.294 & 1.682 \\
& HER + TD3-BC (\textdagger) \cite{fujimoto2021minimalist}& 0 & 0.213 & 10.724 \\
& HER + IQL (\textdagger) \cite{kostrikov2021IQL} & 10 & 1.271 & 3.851 \\
& \ours{} + TD3-BC & 70 & 1.269 & 1.040 \\
& \ours{} + IQL & 90 & 1.030 & 0.759 \\
& \ours{} + DWA & 70 & 1.306 & 1.573 \\
\hline

\multirow{6}{*}{\shortstack{Scenario 3}} 
& DWA (\textdagger) \cite{fox1997dynamic} & 0 & 0.205 & 6.783 \\
& HER + TD3-BC (\textdagger) \cite{fujimoto2021minimalist} & 0 & 0.265 & 7.062 \\
& HER + IQL (\textdagger) \cite{kostrikov2021IQL} & 0 & 0.412 & 5.324 \\
& \ours{} + TD3-BC & 10 & 0.481 & 4.022 \\
& \ours{} + IQL & 80 & 1.136 & 1.263 \\
& \ours{} + DWA & 40 & 0.687 & 4.179 \\

\hline

\end{tabular}
\caption{\small{Performance comparison of different navigation methods across three scenarios. \textbf{Methods with a \textdagger \: have the benefit of using LiDAR}}}
\label{tab:comparisons}
\vspace{-15pt}
\end{table}

Scenario 1 depicts an outdoor environment featuring pedestrians and a crosswalk. In this setting, a human would naturally prefer to avoid pedestrians and navigate along the crosswalk to reach the goal. The MPC planner equipped with the HALO reward model (HALO + DWA) mirrors this human-like behavior, producing smoother, more socially compliant trajectories. As a result, it achieves a lower Fréchet distance and a higher success rate. In contrast, the DWA planner opts for a direct path that ignores the sidewalk and pedestrian context, largely due to its reliance on LiDAR-only perception. While VANP demonstrates the ability to avoid pedestrians, it exhibits jerky motion—particularly within the crosswalk region—attributed to inconsistent action predictions. Similar to the HALO + DWA method, the IQL policy trained with the HALO model (HALO + IQL) also demonstrates social compliance by achieving a low Fréchet distance, although not as low as HALO + DWA. It achieves a success rate comparable to the DWA method, but without relying on LiDAR. The IQL policy trained using the hand-engineered reward (HER + IQL) achieved moderate success rates but did not perform as well as its HALO counterpart. The behavior cloning policy trained with HALO rewards (HALO + BC) was able to successfully avoid obstacles but failed to reach the goal. The BC policy trained using hand-engineered rewards (HER + BC) became stuck in a slight left-hand turn, ultimately driving itself into the bushes.

Scenario 2 features pedestrians and pavement regions under low-light conditions. While the DWA planner avoids pedestrians and proceeds directly toward the goal, both VANP and the MPC planner with the HALO reward model (HALO + DWA) adopt a more socially aware approach by maintaining a large buffer radius around pedestrians—even if it requires adjusting their heading back toward the goal after passing. This behavior reflects the socially compliant navigation induced by the HALO reward model in the MPC-based planner. The IQL and behavior cloning (BC) policies trained with the HALO reward model both achieved relatively high success rates but demonstrated less social compliance. They tended to take more direct paths toward the goal, often passing closer to pedestrians compared to HALO + DWA. The policies trained with the hand-engineered reward (HER) consistently failed to reach the goal. Specifically, the behavior cloning policy trained with the hand-engineered reward (HER + BC) became stuck heading toward the curb.

Scenario 3 takes place in an indoor hallway environment with glass walls and pedestrians. LiDAR-based approaches like DWA struggle in this setting, failing to detect glass walls and resulting in collisions. Similarly, IQL and behavior cloning (BC) methods trained with hand-engineered rewards fail to generalize to this unseen scenario. In contrast, the BC policy trained using the HALO reward demonstrates relatively better performance, indicating improved generalization. Notably, the IQL policy trained with the HALO model exhibits human-like behavior, successfully recognizing both glass walls and pedestrians and navigating the hallway effectively to reach the goal.

\section{Conclusions}
We presented HALO, a novel offline reward learning algorithm that aligns robot navigation with human preferences by leveraging egocentric visual data and expert trajectories. HALO quantifies intuitive human navigation behavior through ranked action preferences and uses the Plackett-Luce model to learn a reward function without requiring hand-engineered heuristics. We demonstrated its adaptability by deploying it in both offline policy learning and model predictive control planners, where HALO achieves superior performance over state-of-the-art baselines across success rate, trajectory safety, and Frechet distance metrics. Our real-world evaluations highlight HALO's robustness across different robot platforms and scenarios.



	





\clearpage

\section*{Limitations}
Our method inherits limitations common to offline reinforcement learning (RL) and learning-based approaches in robotics, particularly in terms of explainability, consistency, and environmental sensitivity.

The lack of explainability both during training and testing has been a limitation of our method. The model performance can be very inconsistent, and when the model fails to avoid an obstacle, it is often unclear why. During the training process, different versions of the model demonstrate certain tendencies, where a checkpoint model may be great at navigating outdoor scenes, and another checkpoint might be better at avoiding people while indoors. The lack of visibility towards model tendencies until testing is a limitation to our methods and many learning-based methods. 

The method has demonstrated sensitivity to lighting conditions, where the robot will fail to avoid obstacles in scenarios where it has successfully navigated before. However, our method has demonstrated capability to navigate at night, or in conditions where there are many reflective surfaces, glass and glares. 

Our method is centered around a reward model that is based on its vision which has a limited field of view. The navigational robot employing our method has demonstrated object impermanence, where the robot may successfully avoid an obstacle only to turn towards it once the obstacle can no longer be seen. In addition, since our method was trained on a variety of robots with varying size, width, and turning radii, the robot trained on our method showed a lack of awareness of its own width. Combined with a narrow field of view, the robot can sometimes avoid an obstacle only to turn back into it. By the time the robot sees an obstacle again it is too late to change course. The method struggles especially when the turning clearance is tight.



\bibliography{example}  

\clearpage
\section{Appendix}

\subsection{Results and Analysis}

This section provides definitions for the evaluation metrics used in the main paper and presents additional qualitative results comparing HALO-based and hand-engineered reward policies in diverse real-world navigation scenarios.

\subsubsection{Metrics}

\label{appdx:metrics}
\begin{table}[H]
\centering
\setlength{\abovecaptionskip}{6pt}
\renewcommand{\arraystretch}{1.4}
\setlength{\belowcaptionskip}{6pt}  
\begin{tabular}{|p{3.5cm}|p{5.2cm}|p{5.3cm}|}
\hline
\textbf{Metric} & \textbf{Definition} & \textbf{Description} \\
\hline
\textbf{Success Rate} &
\[
\frac{N_{\text{success}}}{N_{\text{total}}}
\] &
Fraction of episodes where the robot reaches the goal without collisions. $N_{\text{total}} = 10$\\
\hline
\textbf{Normalized Trajectory Length} &
\[
\frac{L_{\text{actual}}}{L_{\text{expert}}}
\] &
Ratio of the executed trajectory length to the expert trajectory length. Lower is better. \\
\hline
\textbf{Fréchet Distance to \newline Expert Trajectory} &
{\small
\[
\inf_{\alpha, \beta} \max_{t \in [0,1]} \left\| T_r(\alpha(t)) - T_e(\beta(t)) \right\|
\]
}
&
Measures the maximum distance between points on the robot trajectory \(T_r\) and expert trajectory \(T_e\) under continuous, non-decreasing time reparameterizations. This metric captures both spatial proximity and temporal alignment. Lower is better.
 \\
\hline
\end{tabular}
\caption{Evaluation metrics used for assessing navigation performance.}
\label{tab:metrics}
\end{table}

All metrics are averaged over both the successful (reaching the goal) and unsuccessful trials (collision/ not reaching the goal)

\subsubsection{Further Qualitative Analysis}

\label{appdx:qual}
\begin{figure}[H]
  \centering
  \begin{subfigure}{\textwidth}
    \centering
    \includegraphics[width=\textwidth]{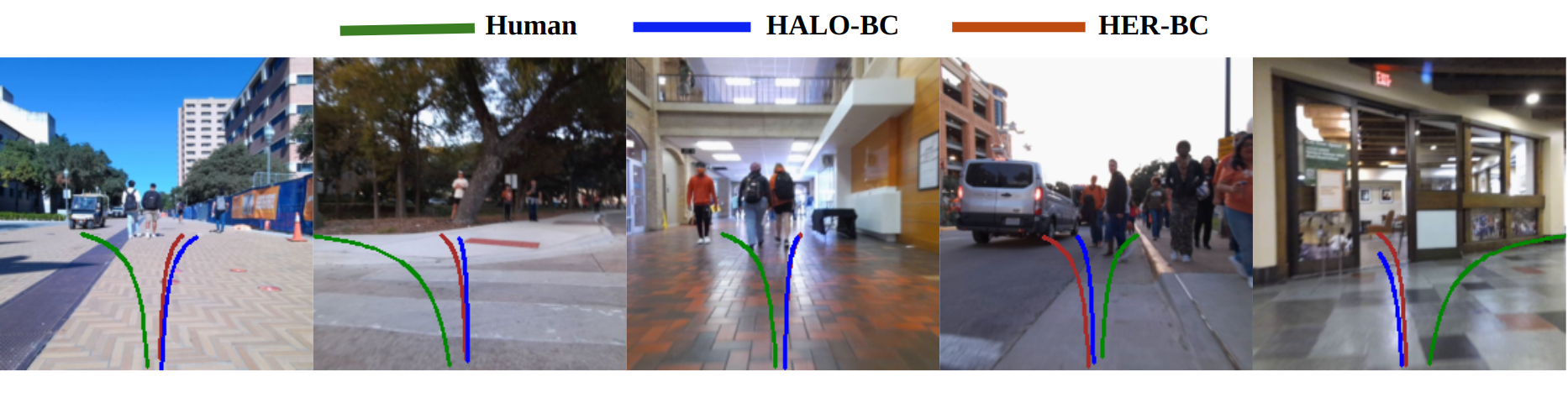}
  \end{subfigure}
  
  \begin{subfigure}{\textwidth}
    \centering
    \includegraphics[width=\textwidth]{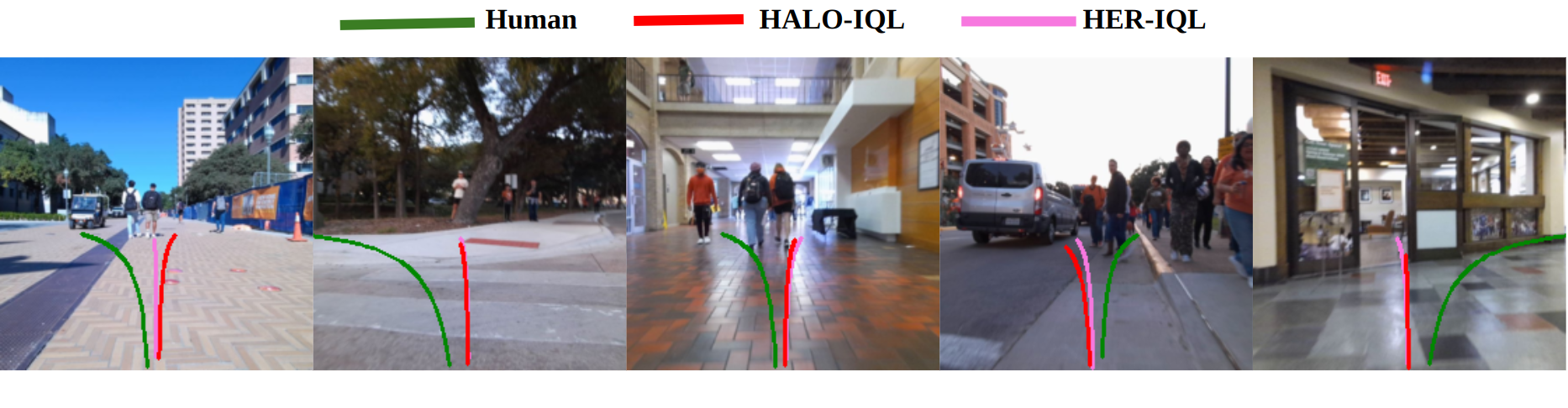}
  \end{subfigure}
  \caption{Qualitative Analysis for Behavioral Cloning (BC) and Implicit Q-Learning (IQL) policies for HALO and Hand Engineered Reward (HER)}
  \label{fig:add-qual}
\end{figure}

(top) illustrates example outputs from HALO-BC and HER-BC policies across varied navigation scenarios. The bottom panel shows similar comparisons for IQL policies.

Figure ~\ref{fig:add-qual} \textbf{(top)} illustrates example outputs from HALO-BC and HER-BC policies across varied navigation scenarios, while Figure~\ref{fig:add-qual} \textbf{(bottom)} presents comparable results for Implicit Q-Learning (IQL) policies. In both cases, we showcase five egocentric images and compare the actions selected by the policies to those taken by a human operator.

The HALO-BC/HALO-IQL policies are trained using reward models learned from human preference feedback, while the hand-engineered reward (HER) BC/IQL policies are trained using manually defined rewards (see Section~\ref{sec:baselines}). In all images, the human-selected actions correspond to higher velocities or sharper turns, while the BC/IQL policies tend to produce more conservative actions. The divergence between human and policy behavior may be attributable to the human operator’s access to contextual information beyond the robot’s egocentric view. For example, in the leftmost image, the human executes a sharp left turn, potentially into oncoming traffic. While this action may appear suboptimal based solely on the visible scene, it could reflect additional cues—such as an approaching agent or a clearer path outside the frame. Given only the egocentric input, the actions chosen by the BC/IQL policies often appear more cautious and contextually appropriate. Each trajectory represents a three-second prediction horizon; the policies will generate new actions before the current trajectory completes.

The HALO-trained policies generate plausible and socially compliant behaviors across diverse scenarios. In the crosswalk image (Figure \ref{fig:add-qual} second from the left), while the human prepares to turn left, the policies prioritize entering the sidewalk, reflecting goal-directed yet safe behavior. When pedestrians are present, the policies prefer to pass on the right, consistent with social norms. In the second-to-last image—a crowded scenario—the human elects to turn right over a yellow curb and through a dense group of pedestrians. This may be feasible due to the use of a legged robot in the SCAND dataset, but the policies instead select a leftward trajectory, trading crowd avoidance for slight proximity to vehicle lanes. In the rightmost image, the policies successfully infer a trajectory through an open door, demonstrating spatial reasoning using only egocentric input.

\subsection{Implementation Details}

\label{appdx:impl_det}
This section details the policy network implementation, the CNN used to process trajectory masks, the key hyperparameters, and model dimensions used in \ours{}.
\paragraph{Policy Network.} 

The policy network shares the same visual encoder and MLP structure as the reward model but replaces trajectory-based attention with uniform aggregation. Specifically, patch embeddings \(F_t\) are averaged to obtain a global context vector, which is passed through an MLP to predict the parameters of a Gaussian distribution over actions. During deployment, actions are sampled from this distribution; during training, gradients are propagated using the reparameterization trick. Q and V networks follow the same structure as the reward model and the policy respectively but output a single scalar value instead of action parameters.

\paragraph{Trajectory Mask CNN.}
To convert the binary trajectory mask \(a_t^m\) into a spatial weighting map \(W_t\), we use a lightweight convolutional neural network with the following structure:

\begin{itemize}
    \item \texttt{Conv2D(1, 4, kernel\_size=3, stride=1, padding=1)}, ReLU
    \item \texttt{MaxPool2D(kernel\_size=2, stride=2)}
    \item \texttt{Conv2D(4, 8, kernel\_size=3, stride=1, padding=1)}, ReLU
    \item \texttt{Conv2D(8, 4, kernel\_size=3, stride=1, padding=1)}
    \item \texttt{Conv2D(4, 1, kernel\_size=3, stride=1, padding=1)}, Sigmoid
\end{itemize}

\subsubsection{Model Parameters}

\begin{table}[H]
\centering
\setlength{\abovecaptionskip}{6pt}

\renewcommand{\arraystretch}{1.2}
\begin{tabular}{lll}
\toprule
\textbf{Symbol} & \textbf{Definition} & \textbf{Value} \\
\midrule
\(H, W\)         & Input image height and width & 224, 224 \\
\(H_m, W_m\)     & Mask resolution after homography projection & 32, 32 \\
\(N_p\)          & Number of image patches (from DINO-v2) & 256 \\
\(d_e\)          & Embedding dimension of each patch & 384 \\
\(f_t\)          & Aggregated feature vector from weighted patches & \(\mathbb{R}^{384}\) \\
\(R_t\)          & Scalar reward output & \(\mathbb{R}\) \\
\(n_v, n_\omega\) & Number of discrete velocity and angular bins & 5, 5 \\
& Dropout rate for MLPs & 0.1 \\

\bottomrule
\end{tabular}
\caption{\small{Definitions and values of key model dimensions and constants used throughout the architecture.}}
\vspace{-15pt}
\label{tab:model_params}
\end{table}

\subsection{Training Methodology}

\subsubsection{Reward Model Training}

The reward model is trained using the Plackett-Luce (PL) loss, which aligns predicted scalar rewards with human-labeled preference rankings over discrete action sets. We apply a 20\% validation split and employ early stopping based on validation loss. In addition to the PL loss, we incorporate three auxiliary loss terms to regularize the reward predictions:

\begin{itemize}
    \item \textbf{Reward Diversity Regularization:} Encourages the model to assign more distinct rewards to dissimilar actions. This is achieved by penalizing pairs of actions whose reward differences are too small relative to their distances in action space, using a symmetric hinge-based loss:
    \[
    \mathcal{L}_{\text{div}} = \mathbb{E}_{(a_i, a_j)} \left[ \max\left(0, \|a_i - a_j\| - \frac{1}{c} |r_i - r_j| \right) + \max\left(0, c |r_i - r_j| - \|a_i - a_j\| \right) \right]
    \]
    where \(a_i, a_j\) are actions and \(r_i, r_j\) are their predicted rewards. The constant \(c\) defines the desired proportionality between action and reward differences.

    \item \textbf{Focal Regression Regularization:} A weighted mean squared error (MSE) that penalizes large reward errors more heavily, thereby encouraging the model to capture both order (from PL loss) and relative magnitudes:
    \[
    \mathcal{L}_{\text{focal}} = \mathbb{E} \left[ \left( \hat{r} - r \right)^2 \cdot \left| \hat{r} - r \right|^\gamma \right]
    \]
    where \( \hat{r} \in \mathbb{R}^Q \) and \( r \in \mathbb{R}^Q \) are the predicted and target reward vectors, and \(\gamma\) controls the emphasis on larger errors.
    
    \item \textbf{L2 Regularization:} A regularization term that penalizes large reward magnitudes to improve numerical stability and prevent overly confident predictions:
    \[
    \mathcal{L}_{\text{L2}} = \mathbb{E} \left[ \| \hat{r} \|_2^2 \right]
    \]
    where \( \hat{r} \) denotes the vector of predicted rewards.

\end{itemize}

The final training objective is given by:
\[
\mathcal{L}_{\text{total}} = \mathcal{L}_{\text{PL}} + \lambda_1 \mathcal{L}_{\text{div}} + \lambda_2 \mathcal{L}_{\text{L2}} + \lambda_3 \mathcal{L}_{\text{focal}}
\]

\begin{figure}[htpb]
    \centering
    \includegraphics[width=0.32\textwidth]{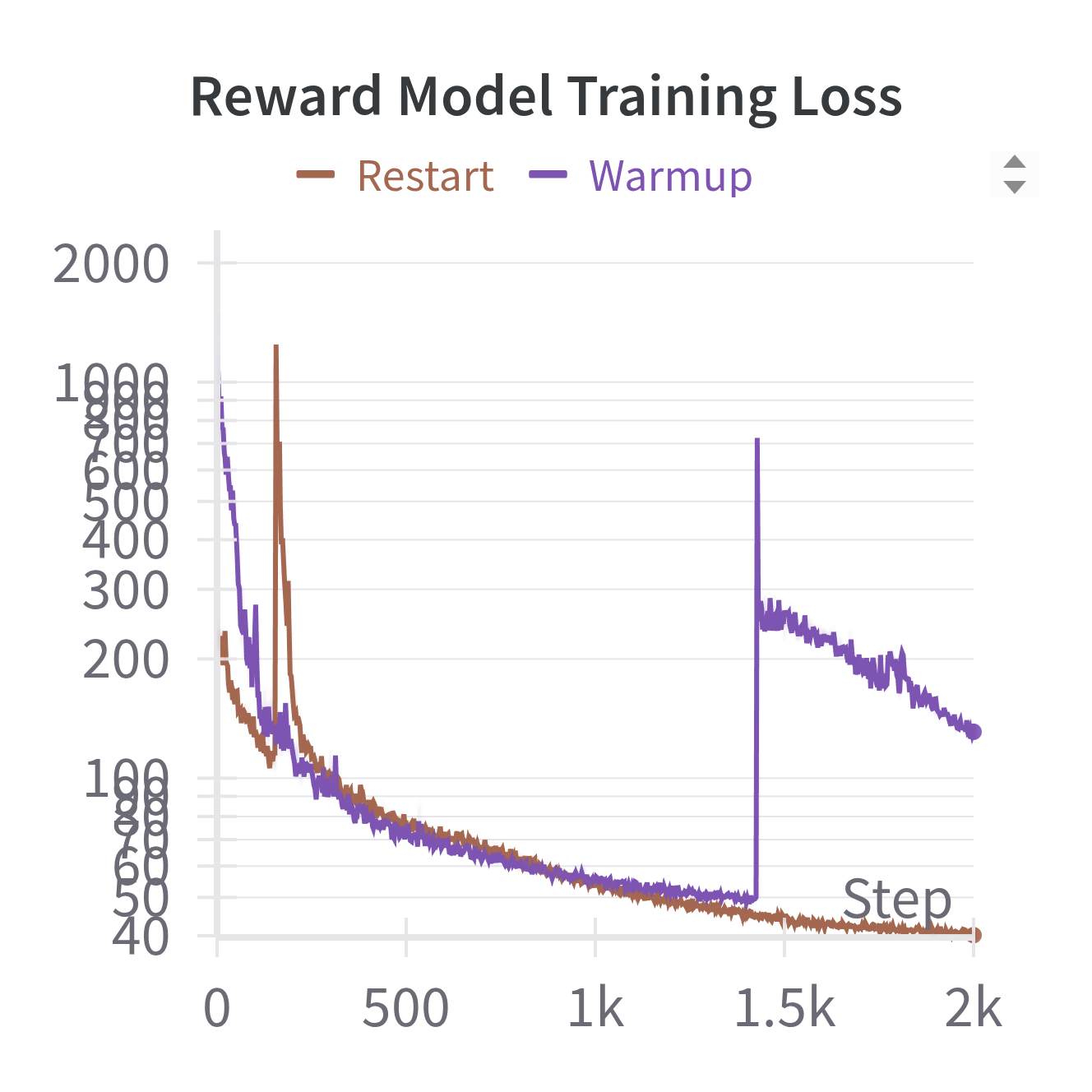}
    \includegraphics[width=0.32\textwidth]{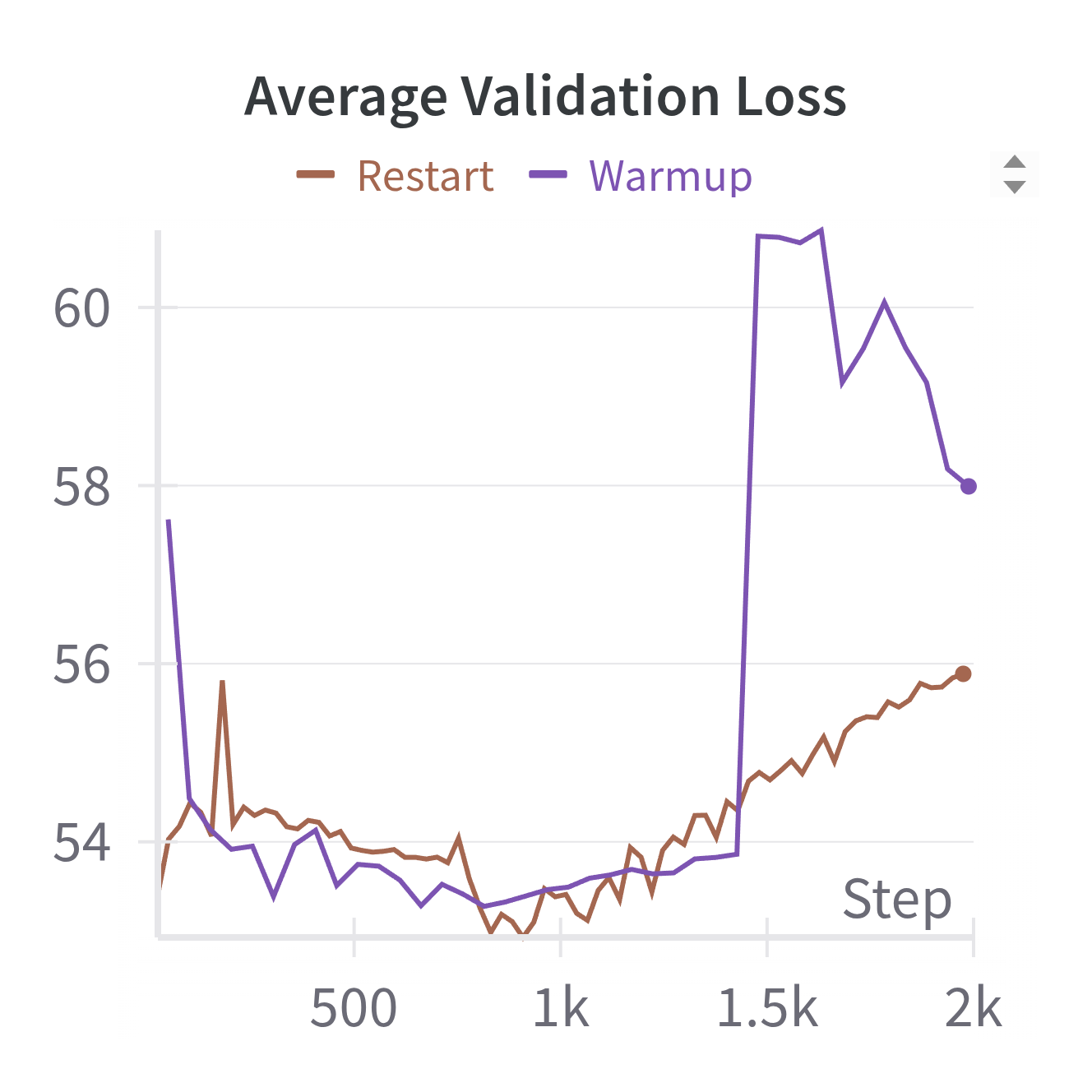}
    \includegraphics[width=0.32\textwidth]{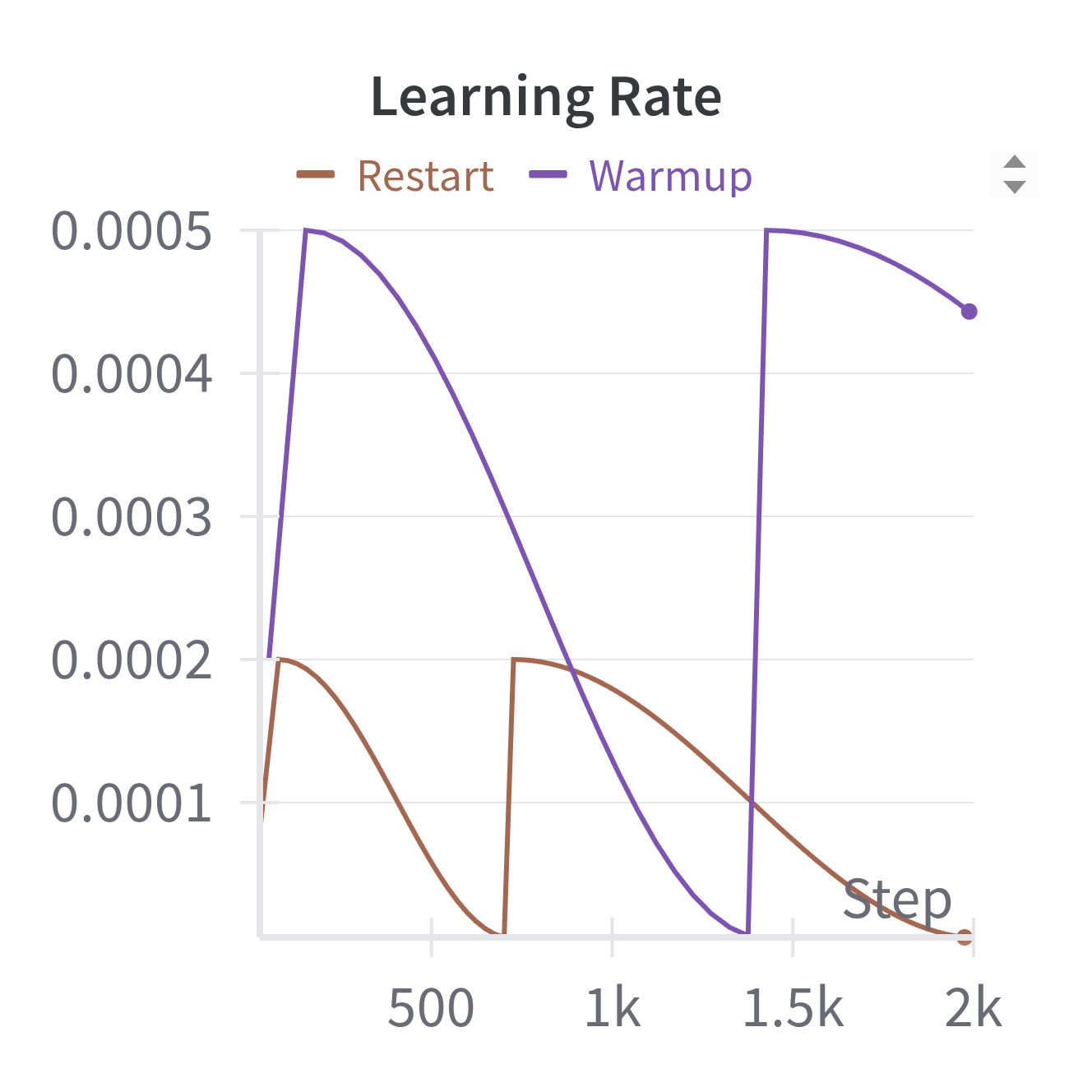}
    \caption{Reward Model Training Loss Curves: }
    \label{fig:rm_training_curves}
    \vspace{-15pt}

\end{figure}

The training history of the reward model is shown in figure \ref{fig:rm_training_curves}.  We employed a two-stage training strategy: an initial warm-up phase with a higher learning rate, followed by a restarted training phase with a lower learning rate. Training was terminated early once signs of overfitting appeared during the restart phase. As noted in the hyperparameter table \ref{tab:reward_model_hparams}, the cosine-annealing with warm-restarts scheduler was used, which has resulted in spikes in the loss curves. 

\begin{table}[h!]
\centering
\setlength{\abovecaptionskip}{6pt}
\begin{tabular}{ll}
\toprule
\textbf{Hyperparameter} & \textbf{Value} \\
\midrule
Optimizer & AdamW \\
Learning rate & \(2 \times 10^{-4}\) \\
Weight decay & \(1 \times 10^{-3}\) \\
Batch size & 256 \\
Number of epochs & 200 \\
Scheduler & Cosine Annealing with Warm Restarts \\
Warm-up epochs & 3 \\
Cosine \(T_0\) & 25 \\
Cosine \(T_{\text{mult}}\) & 2 \\
Validation split & 20\% \\
\midrule
\(\lambda_1\) (reward diversity) & 2.0 \\
\(\lambda_2\) (L2 penalty) & 0.01 \\
\(\lambda_3\) (focal regression) & 0.05 \\
\bottomrule
\end{tabular}
\caption{\small{Training hyperparameters used for reward model optimization.}}
\label{tab:reward_model_hparams}
\vspace{-15pt}
\end{table}

\subsubsection{Policy Training}

We train goal-conditioned policies using two offline methods: Behavior Cloning (TD3-BC) \cite{fujimoto2018addressing}, and Implicit Q-Learning (IQL) \cite{kostrikov2021IQL}. Since the emphasis of our method is the training of the reward model, we leveraged the standard implementation of the clean offline reinforcement learning (CORL) \cite{tarasov2022corl} library. The offline RL implementations such as learning rate, loss functions, etc were mostly kept as-is with changes implemented when appropriate, such as including image transformations for the DINOV2 encoder, and providing the trajectory mask to the critic network. In all cases, the reward signal is provided by our trained preference model.  Similar to the reward model, pytorch's CosineAnnealingWarmRestarts option was used for training the policy as well. This has contributed to the cyclic rise and fall in the loss curves.

\begin{figure}[htpb]
    \centering
    \includegraphics[width=0.32\textwidth]{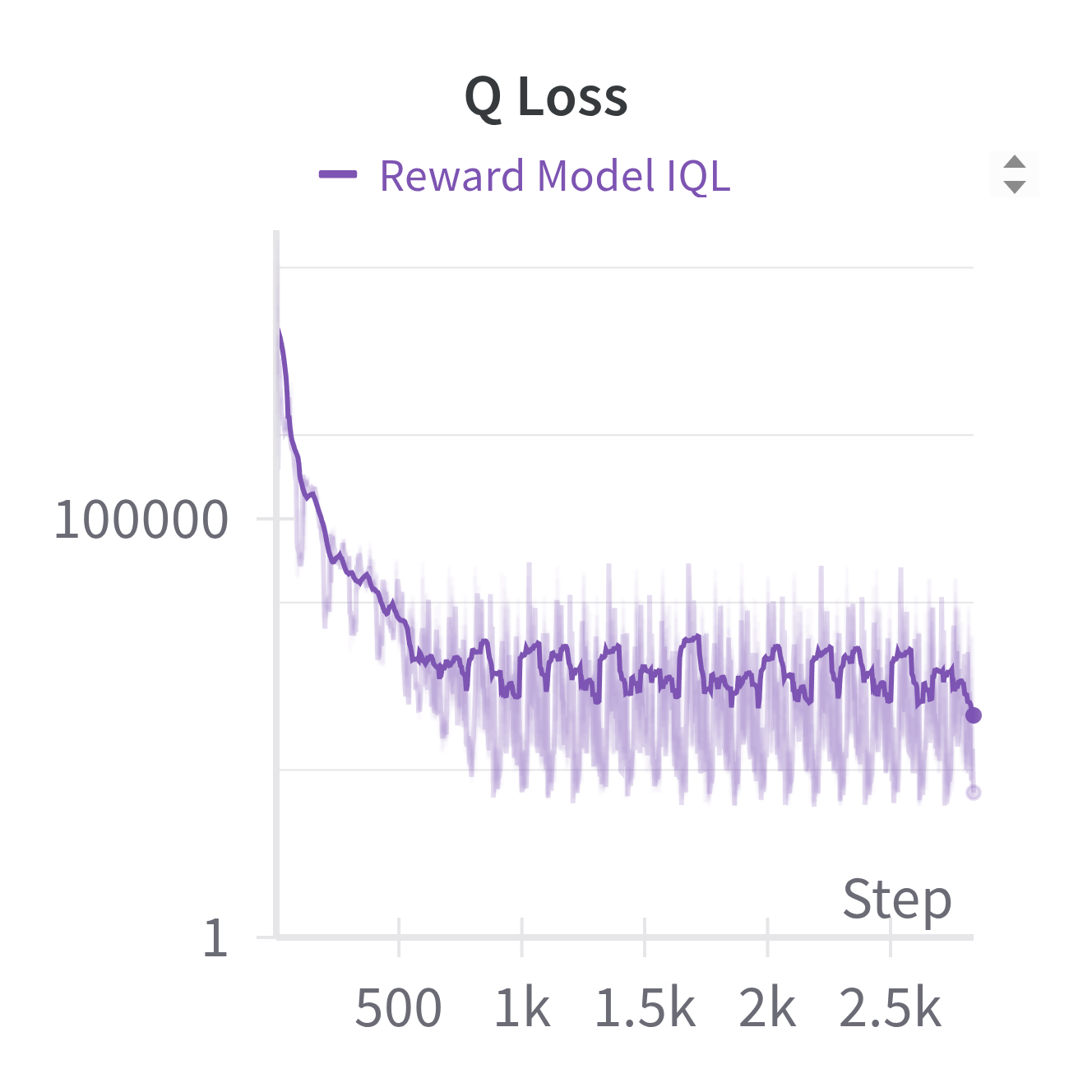}
    \includegraphics[width=0.32\textwidth]{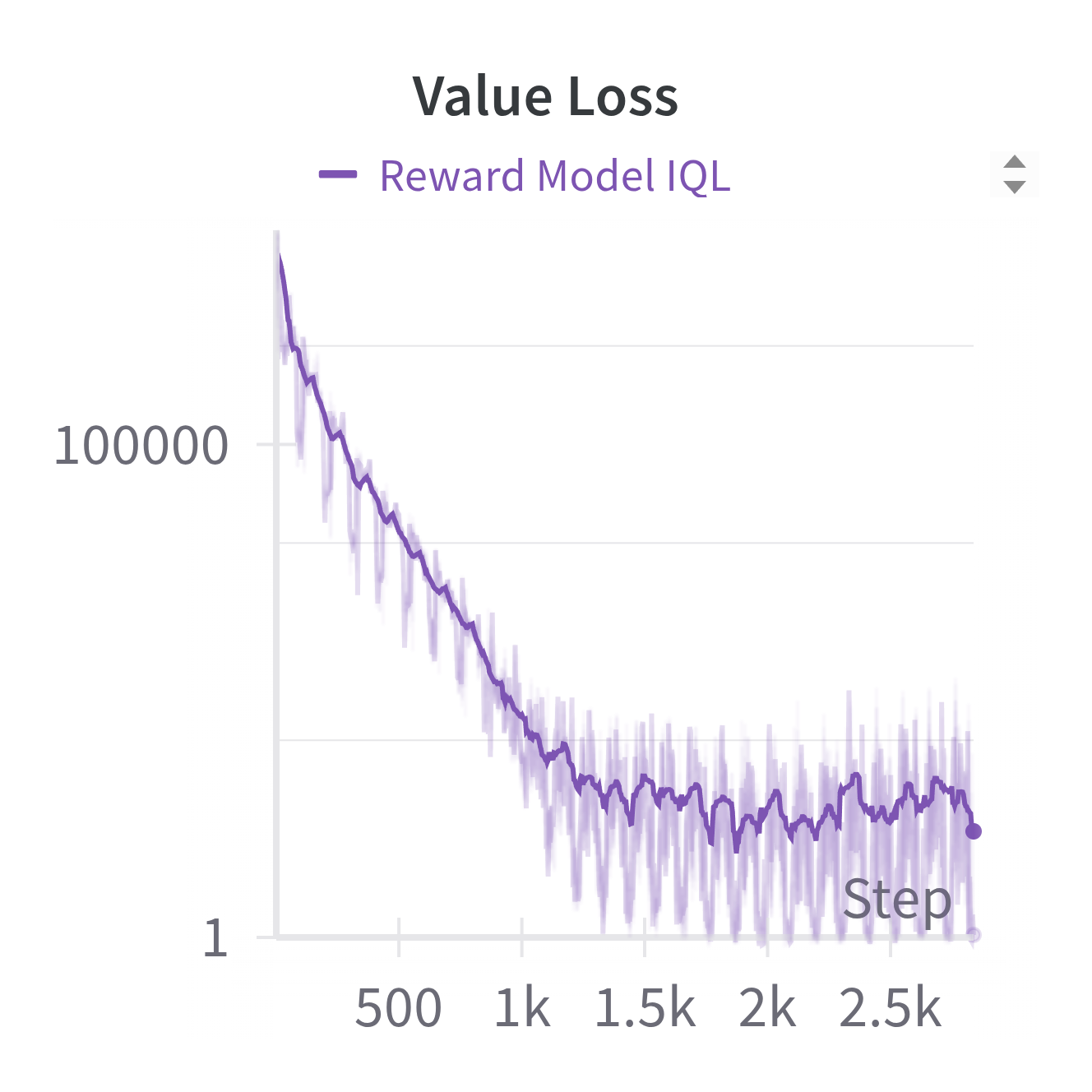}
    \includegraphics[width=0.32\textwidth]{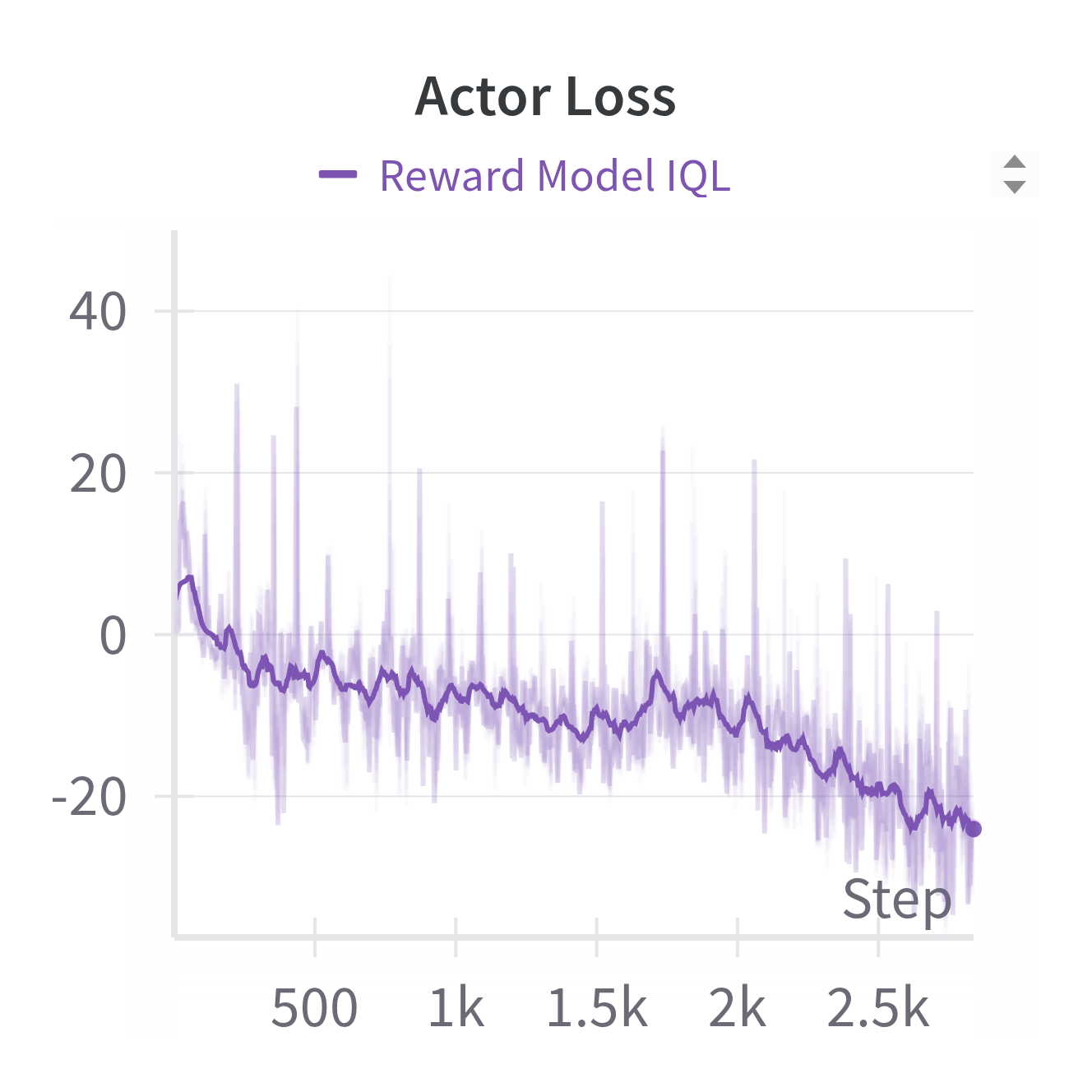}
    \caption{Training loss curves for IQL training}
    \label{fig:iql_training_curves}
\end{figure}

\begin{figure}[htpb]
    \centering
    \includegraphics[width=0.8\textwidth]{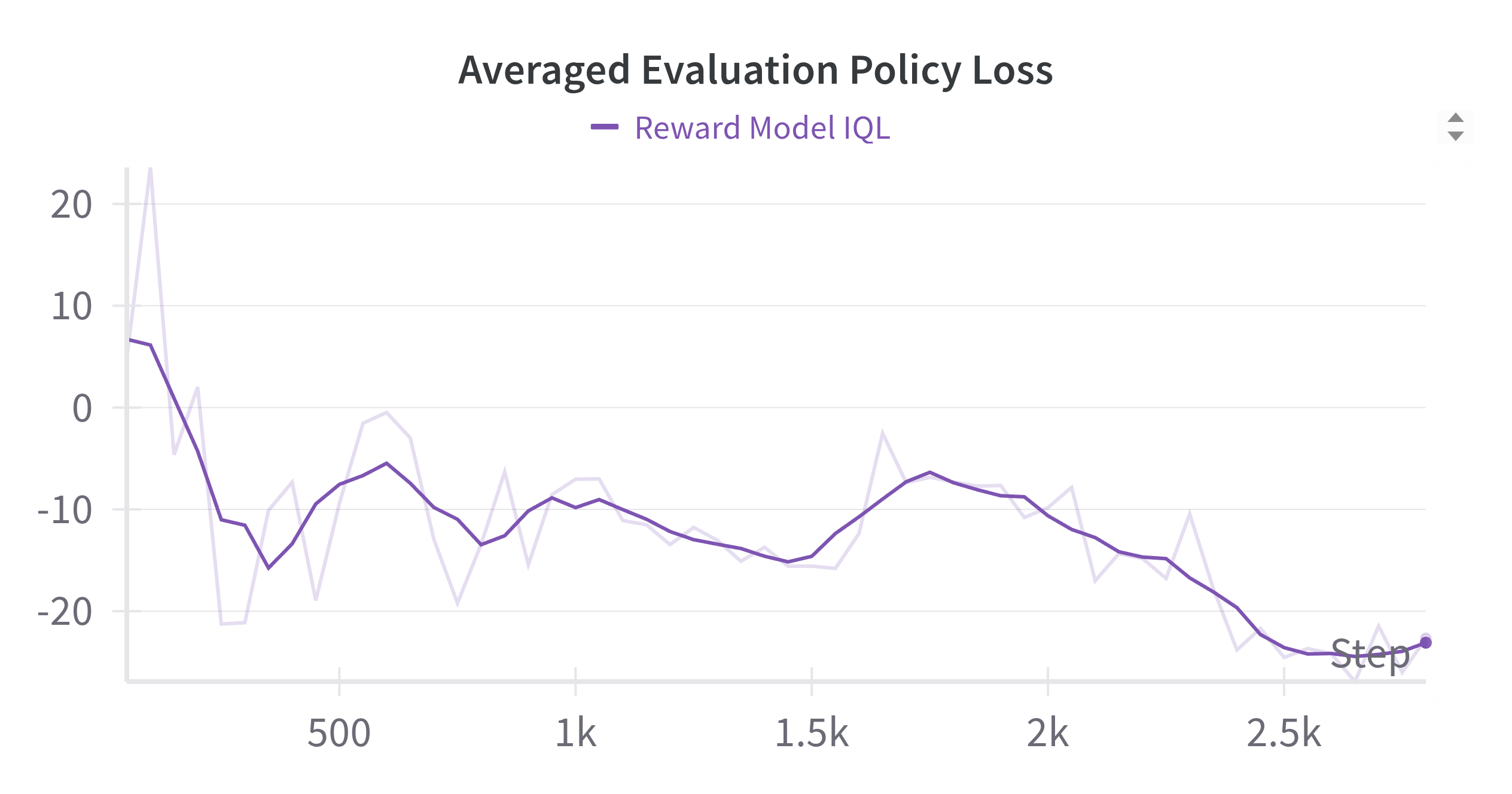}
    \caption{Evaluation curve for IQL training}
    \label{fig:iql_eval_curves}
\end{figure}

The loss curves for the IQL policy trained with our custom reward model are shown in figures \ref{fig:iql_training_curves} and \ref{fig:iql_eval_curves}. The overlay of the loss shown is a rolling average over 20 steps, while the actual loss values are faded. The overlay for the policy evaluation loss over a holdout set was averaged over only 5 steps, since this evaluation loss was already an average over evaluating the validation set over multiple batches.

\begin{figure}[htpb]
    \centering
    \includegraphics[width=0.45\textwidth]{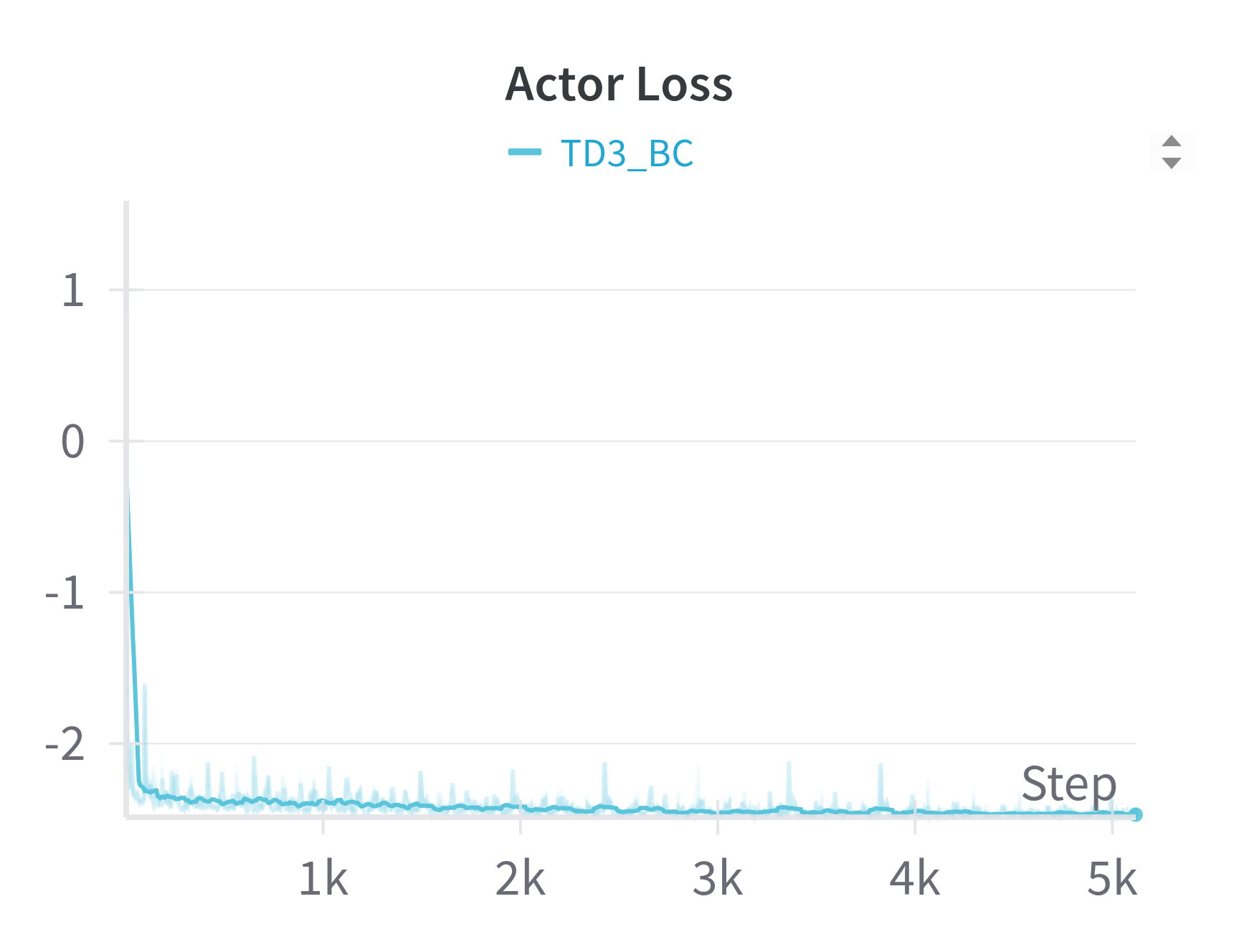}
    \includegraphics[width=0.45\textwidth]{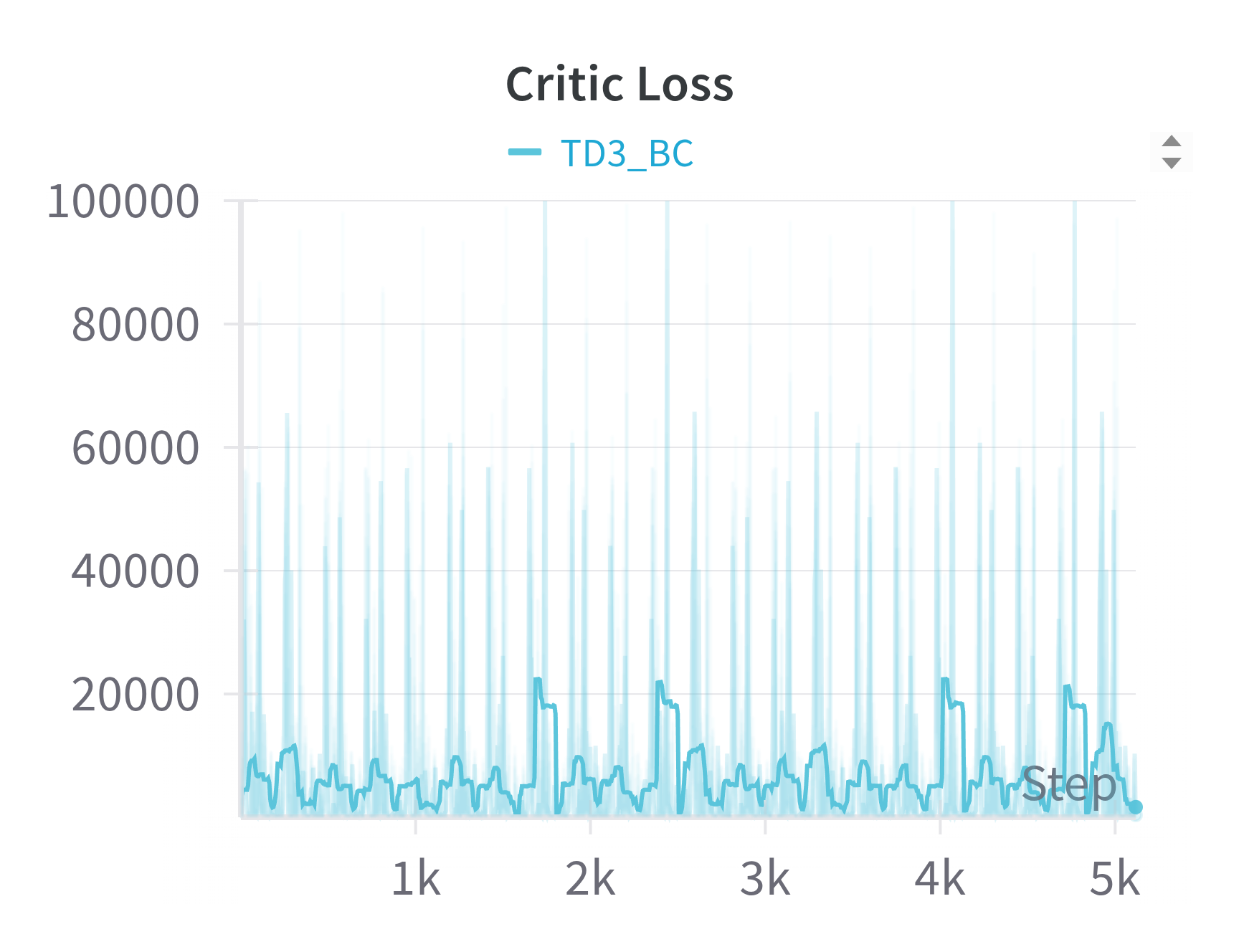}
    \caption{Training loss curves for TD3-BC}
    \label{fig:bc_training_curves}
\end{figure}

\begin{figure}[htpb]
    \centering
    \includegraphics[width=0.45\textwidth]{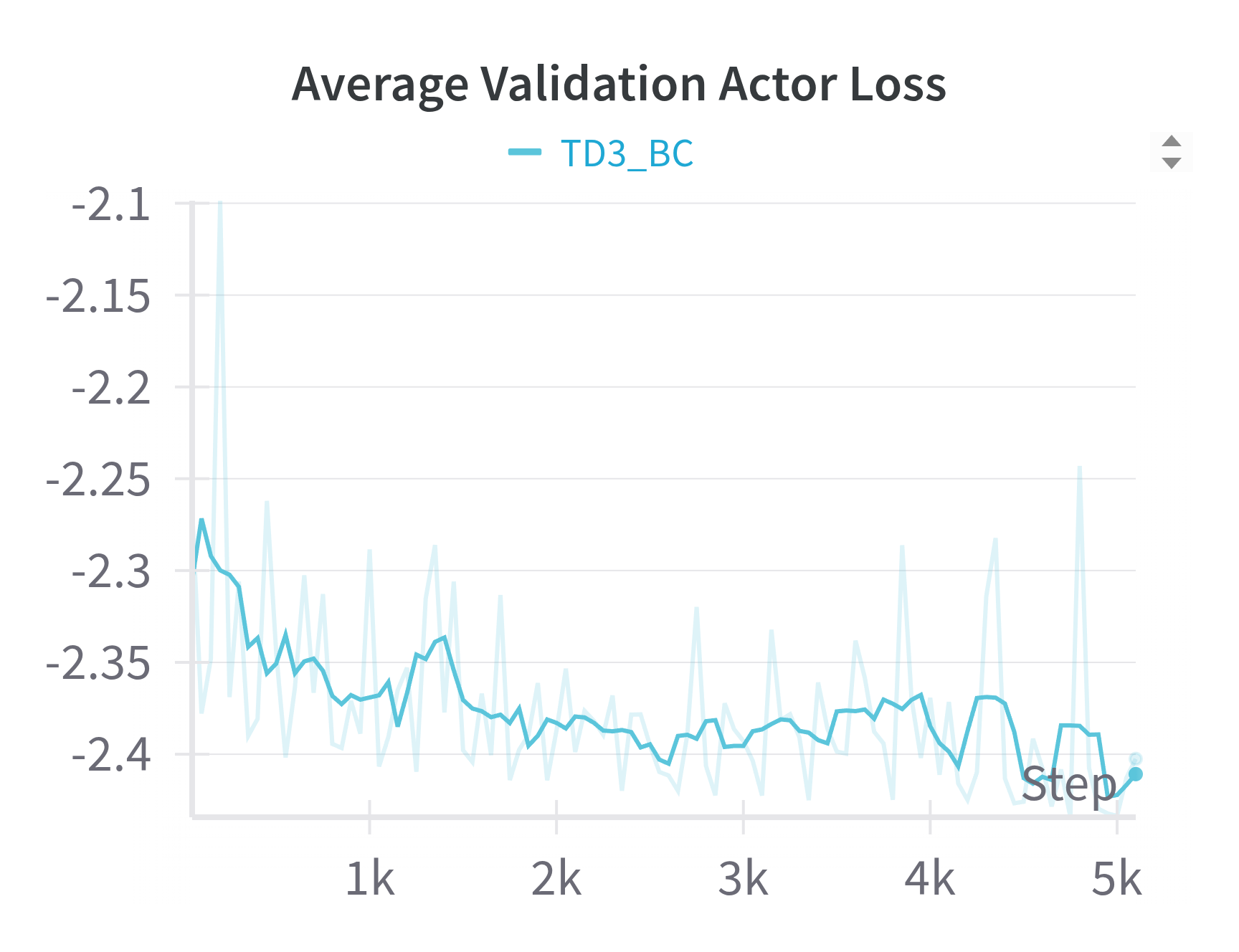}
    \includegraphics[width=0.45\textwidth]{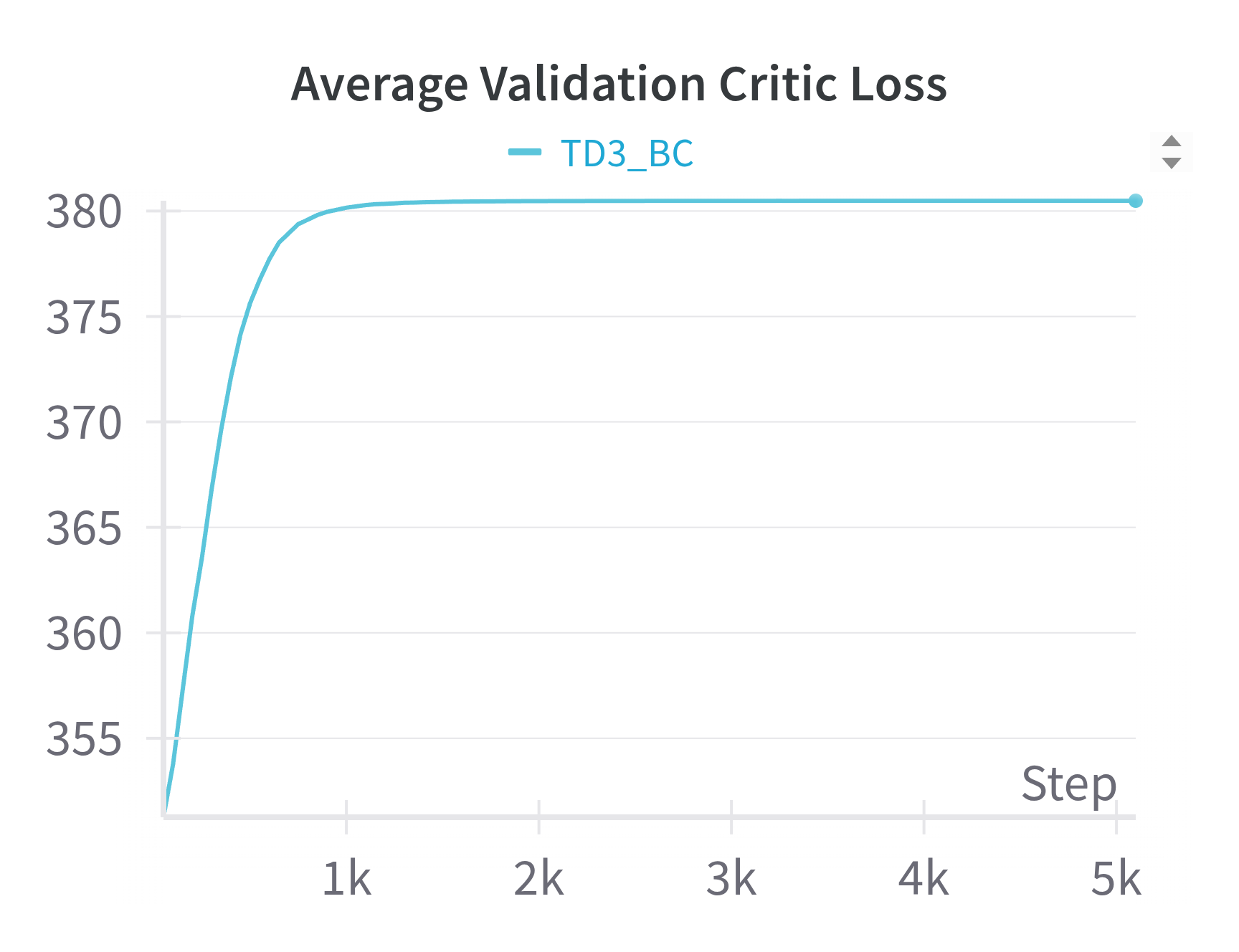}
    \caption{Evaluation loss curves for TD3-BC}
    \label{fig:bc_eval_curves}
\end{figure}

Similarly, the TD3-BC are shown in figures \ref{fig:bc_training_curves} and \ref{fig:bc_eval_curves}. The rolling average over 20 steps was overlaid for the critic loss, and the overlay for the validation actor loss was averaged over 5 steps. The learning rate for TD3-BC was kept constant at 0.0001.

.
\subsection{Baselines and Classical Control Details}
\label{sec:baselines}

\subsubsection{Hand-Engineered Reward for Offline Comparison Methods}

Each action \((v, \omega)\) is simulated forward in time into a trajectory \(\tau_{v,\omega} = \{(x_t, y_t, \theta_t)\}_{t=1}^T\). \textit{The comparison methods benefit from access to LiDAR, enabling the use of hand-crafted reward functions that exploit precise geometric cues for obstacle avoidance.}

The reward is computed as a weighted sum of three cost components:

\[
R(v, \omega) = - \left( 
\lambda_{\text{goal}} \cdot \underbrace{d_{\text{goal}}(\tau_{v,\omega}, g)}_{\text{goal distance}} \:
+ \: \lambda_{\text{head}} \cdot \underbrace{\theta_{\text{err}}(\tau_{v,\omega}, g)}_{\text{heading error}} \:
+ \: \lambda_{\text{obs}} \cdot \underbrace{c_{\text{obs}}(\tau_{v,\omega})}_{\text{obstacle cost}} \:
+ \: \lambda_{\text{smooth}} \cdot \underbrace{\|a - a_{\text{prev}}\|_2}_{\text{smoothness}} 
\right)
\]

where:
\begin{itemize}
    \item \(d_{\text{goal}}(\tau_{v,\omega}, g)\): Euclidean distance between the final state of the trajectory and the goal position,
    \item \(\theta_{\text{err}}(\tau_{v,\omega}, g)\): absolute difference in heading between the final trajectory angle and the goal orientation,
    \item \(c_{\text{obs}}(\tau_{v,\omega})\): an exponential penalty that emphasizes proximity, assigning higher cost to trajectories near LIDAR-detected obstacles by weighting inversely with distance.
    \item \(\|a - a_{\text{prev}}\|_2\): L2 norm enforcing smoothness relative to the previous action.
\end{itemize}

We use the following weights, empirically selected based on our offline dataset:
\[
\lambda_{\text{goal}} = \frac{1}{5}, \quad
\lambda_{\text{head}} = 1.0, \quad
\lambda_{\text{obs}} = \frac{1}{8}, \quad
\lambda_{\text{smooth}} = 1.0
\]
\subsubsection{Goal Conditioning with \ours{}}

To enable goal-directed navigation with our learned reward model, we modify the hand-engineered reward by replacing the obstacle penalty with the output of \ours{}, which captures human preferences, including implicit notions of obstacle avoidance and social compliance:

\[
R(v, \omega) = - \left( 
\lambda_{\text{goal}} \cdot d_{\text{goal}}(\tau_{v,\omega}, g) 
+ \lambda_{\text{head}} \cdot \theta_{\text{err}}(\tau_{v,\omega}, g) 
\right) 
+ \lambda_{\text{halo}} \cdot \underbrace{R_{\text{\ours{}}}(s, a)}_{\text{learned reward}}
\]

This formulation preserves geometric alignment to the goal while incorporating semantics from human-labeled preferences. All existing weights are retained from the hand-engineered formulation, and we set $\lambda_{\text{halo}} = \frac{1}{5}$.

\subsubsection{DWA Planner Configuration}

We adopt a standard goal-conditioned Dynamic Window Approach (DWA) planner with a 1.0-second planning horizon, 5 Hz control frequency, and 25 uniformly sampled linear and angular velocities. To integrate learned preferences, we simply add the negative of our reward model's output as a cost term:

\[
J(a) = J_{\text{DWA}}(s, a) - \lambda_{\text{halo}} \cdot R_{\text{\ours{}}}(s, a)
\]

Here, \(J_{\text{DWA}}(s, a)\) denotes the hand-crafted reward based on goal distance, heading error , and action transition smoothing (excluding obstacle cost), and \(R_{\text{\ours{}}}(s, a)\) is the learned preference-based reward. We set \(\lambda_{\text{halo}} = \frac{1}{5}\).

\end{document}